# A Hybrid Approach to Measure Semantic Relatedness in Biomedical Concepts

Katikapalli Subramanyam Kalyan and Sivanesan Sangeetha

Department of Computer Applications, NIT Trichy, India-620015.

kalyan.ks@yahoo.com, sangeetha@nitt.edu

**ABSTRACT**

**Objective:** This work aimed to demonstrate the effectiveness of a hybrid approach based on Sentence BERT model and retrofitting algorithm to compute relatedness between any two biomedical concepts.

**Materials and Methods:** We generated concept vectors by encoding concept preferred terms using ELMo, BERT, and Sentence BERT models. We used BioELMo and Clinical ELMo. We used Ontology Knowledge Free (OKF) models like PubMedBERT, BioBERT, BioClinicalBERT, and Ontology Knowledge Injected (OKI) models like SapBERT, CoderBERT, KbBERT, and UmlsBERT. We trained all the BERT models using Siamese network on SNLI and STSb datasets to allow the models to learn more semantic information at the phrase or sentence level so that they can represent multi-word concepts better. Finally, to inject ontology relationship knowledge into concept vectors, we used retrofitting algorithm and concepts from various UMLS relationships. We evaluated our hybrid approach on four publicly available datasets which also includes the recently released EHR-RelB dataset. EHR-RelB is the largest publicly available relatedness dataset in which 89% of terms are multi-word which makes it more challenging.

**Results:** Sentence BERT models mostly outperformed corresponding BERT models. The concept vectors generated using the Sentence BERT model based on SapBERT and retrofitted using UMLS-related concepts achieved the best results on all four datasets.

**Conclusions:** Sentence BERT models are more effective compared to BERT models in computing relatedness scores in most of the cases. Injecting ontology knowledge into concept vectors further enhances their quality and contributes to better relatedness scores.

**INTRODUCTION**

Semantic relatedness is one of the fundamental research topics in natural language processing [1]. Semantic relatedness represents how close two concepts are i.e., how much semantic linkage exists between them. For example, Low-Density Lipoprotein (LDL) and Zocor are semantically related because Zocor is used to reduce LDL in the blood. Research in semantic relatedness involves developing novel methods to quantify relatedness in a way that the calculated relatedness scores are close to human judgments. Semantic similarity is a fine-grained version of semantic relatedness where the two concepts share more common properties [1]. For example, Lipitor and Zocor are semantically similar because both are drugs, and both are used to reduce LDL in the blood. Semantically related concepts need not be semantically similar. For example, Zocor and LDL are semantically related because Zocor is used to reduce LDL in the blood. However, these two are not semantically similar because they don't share more common properties like Lipitor and Zocor. Semantic relatedness measures are useful in many applications. Some of the examples are sentiment analysis [2], plagiarism detection [3], spell correction [4] in the general domain and ontology development [5], information retrieval in electronic health records [6], identification of protein-protein interaction [7] in the biomedical domain.

We can categorize methods for automatic computation of relatedness scores into knowledge-based, distributional, and hybrid depending on the source they leverage. Knowledge-based approaches make use of expert-curated knowledge available in hierarchical knowledge sources. For example, WordNet [9] in general English and Unified Medical Language System (UMLS) [10] in the biomedical domain. Some of the examples of knowledge-based approaches are path-bath measures [11-15], information-content-based measures [16-17], and gloss-based measures [18-19]. The main advantage of knowledge-based approaches is their simplicity. However, developing hierarchical knowledge sources is expensive as well as requires a lot of effort. Moreover, it is not guaranteed that related concepts always have a link in between them which results in poor performance in the case of concepts that are related but not linked in the ontology [20]. Distributional methods involve generating word or Concept Unique Identifier (CUI) vectors from a large text corpus. The main idea behind distributional methods is the distributional hypothesis [21-22] which states that semantics of a word or a CUI is determined by its context. Traditional approaches [8,23] are based on cosine similarity between high dimensional sparse vectors generated using co-occurrence information in the large text corpus. Recent approaches [24-26] are based on low-dimensional dense vectors generated using models like Word2Vec [27] and FastText [28]. Some of the approaches [29-34] generate CUI vectors from a large text corpus. These approaches initially generate sequences of CUIs from text corpus by identifying and normalizing concept mentions and then generate CUI vectors. The main drawback in these approaches is that embeddings are available only for the concepts occurring in the training corpus. Moreover, the quality of concept embeddings for less frequently occurring concepts in the training corpus is limited.

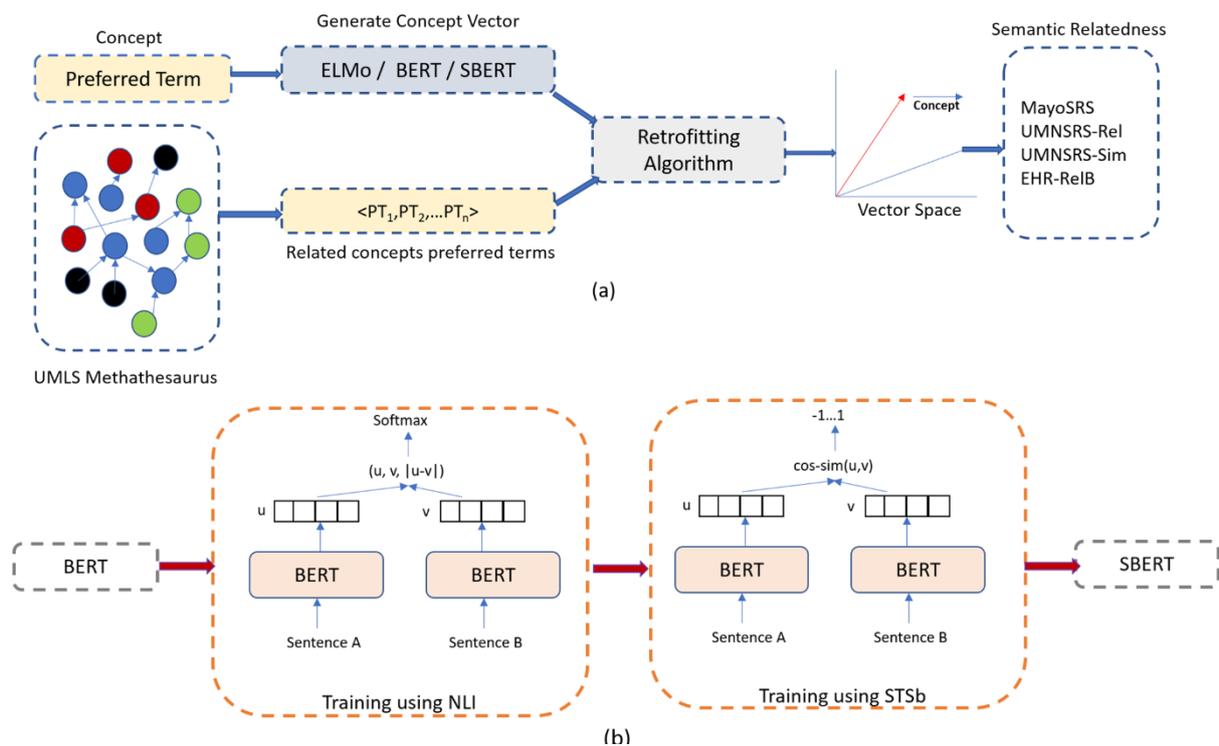

**Figure 1.** (a) Overview of our approach. (b) Training BERT model using Siamese Network on NLI and STSb datasets to get Sentence BERT (SBERT).

A hybrid approach is a combination of knowledge-based and distributional approaches by leveraging knowledge from both ontology and text corpora. Yu et al. [31] proposed a hybrid measure based on CUI embeddings generated using Word2Vec and retrofitting algorithm. However, this method can only be applied to concepts available in the training corpus which limits its application. Recently, Mao and Fung [20] proposed a hybrid relatedness measure based on biomedical word embeddings and graph embeddings. Concept vectors based on word embeddings are not effective in the case of multi-word concepts [6]. Further, leveraging ontology relationship knowledge in the form of graph embeddings is not effective as related concepts need not be always linked in the ontology and due to the limit on the size of the graph, concepts from one or two relationships only can be leveraged. To overcome the drawbacks in the existing hybrid approaches, we propose a hybrid relatedness measure based on domain-specific Sentence BERT model and retrofitting algorithm as shown in Figure 1(a). We strongly believe that this is the first work to generate and explore the use of domain-specific Sentence BERT model for biomedical concept relatedness.

**Background**

Word Embedding Models

Machine Learning or Deep Learning based NLP systems expect numerical input. Traditional approaches convert text to vectors of numbers based on statistical measures and these vectors are sparse, high dimensional, and ignore semantic information. Word embeddings evolved as a much better alternative to these traditional approaches and these are dense, low dimensional, and also encode semantic information. The first generation of embedding models include Word2Vec [27], Glove [35] and FastText [28]. These models are based on shallow neural networks and generate context-independent word embeddings. The second generation of embedding models are context-dependent and include models like ELMo [36], BERT [37], etc.

The performance of embeddings trained on general corpora in biomedical tasks is limited as domain-specific text includes a lot of domain-specific terms. So, biomedical NLP researchers trained embedding models on domain-specific corpora and released those models. For example, BioELMo [38] is trained on PubMed abstracts while ClinicalELMo [39] is trained on MIMIC-III [40] dataset and

SNOMED-CT [41] related Wikipedia pages text. BioBERT [42] is initialized from the general BERT model and further pre-trained on 1M PubMed abstracts. BioClinicalBERT [43] is initialized from the BioBERT model and further pre-trained on the MIMIC-III dataset. Here BioBERT and Bioclinical BERT models are initialized from other BERT models and then further pre-trained on domain-specific corpora. Two versions of PubMedBERT [44] are available a) PubMedBERT trained on PubMed abstracts only and b) PubMedBERT trained on PubMed abstracts and PMC full text. In this paper, we used the PubMedBERT model trained on both abstracts and full text. These models encode only the knowledge available in the training corpora and lack valuable expert knowledge available in domain-specific ontologies like UMLS.

Recently researchers started to focus on leveraging UMLS knowledge to further enhance the domain knowledge encoded in domain-specific embedding models. SapBERT [45] is a BERT-based model obtained by further pre-training PubMedBERT on UMLS using a self-alignment objective. Self-alignment objective using metric learning loss allows the model to cluster the terms which represent the same concept. CoderBERT [46] is also a BERT-based model obtained by further pre-training BioBERT on UMLS synonyms and relations. UmlsBERT [47] is initialized from BioClinicalBERT and further pre-trained using UMLS data. Here masked language modeling objective using multi-label loss allows the model to understand the connections between all the terms under one CUI. Semantic group embedding which is used along with position, word, and segment embeddings leverages UMLS semantic group knowledge. KbBERT [48] is also initialized from BioBERT and further pre-trained on UMLS relation data using triplet classification loss.

Sentence Embedding Models

Sentence embedding models project variable-length text like phrases, sentences to low dimensional vector space in a way that semantically similar texts are closer. Sentence embeddings find applications in various tasks like retrieving relevant sentences, clustering, paraphrase mining [49], and even in domain-specific applications like medical concept normalization [50-51]. Research in learning sentence representations started with Skip-thought vectors [52] followed by Sent2Vec [53], InferSent [54], Universal Sentence Encoder [55] and recently SBERT [49]. Skip-thought is an extension of the skip-gram model [27] to sentences and Sent2vec is an extension of the CBOW model [27] to sentences. InferSent is based on BiLSTM and trained in a supervised manner on both SNLI [55] and MNLI [56] datasets. Universal Sentence Encoder uses transformer encoder and is trained on unlabelled web crawled data followed by training on SNLI dataset. Recently, SBERT has come into the picture which is a BERT-based sentence embedding model trained in a Siamese manner on SNLI, MNLI, and STSb [57] datasets. BERT model is trained on SNLI and MNLI datasets in a Siamese manner with classification objective using the cross-entropy loss and then trained on STSb dataset with regression objective using the mean-squared loss to get SBERT model (refer Figure 1(b)).

Retrofitting Algorithm

Vector representations learned from unlabelled corpora encode only the distributional information available in text corpora. The quality of these embeddings can be further enhanced by injecting ontology knowledge. To inject ontology knowledge into embeddings, Faruqui et al. [58] proposed a simple graph-based method based on belief propagation. This method infers retrofitted embeddings from original embeddings by iteratively reducing the distance between the linked words or concepts i.e., reduces the distance between the retrofitted vector and all its neighbors as well as the distance the between retrofitted vector and the original vector. The advantages of this approach are, it is fast and can be applied to vectors generated by any embedding model.

**MATERIALS AND METHODS**

**Materials**

Unified Medical Language System (UMLS)

UMLS is a large collection of medical terms from over 150 biomedical vocabularies. The contents in UMLS are organized into Metathesaurus, Semantic Network, and Specialist Lexicon. In UMLS, medical terms from different biomedical vocabularies that represent the same clinical concept are mapped to one concept which is assigned with Concept Unique Identifier (CUI). For each CUI, one of the medical terms is chosen as Preferred Term and the rest are referred to as synonym terms. Further, each CUI is assigned with one or more semantic types. The linking between concepts is represented using one of the UMLS relationships as shown in Figure 2. Metathesaurus is the heart of UMLS containing concepts and their associated synonyms in MRCONSO file, related CUIs in MRREL file, CUI definitions in MRDEF file, etc.

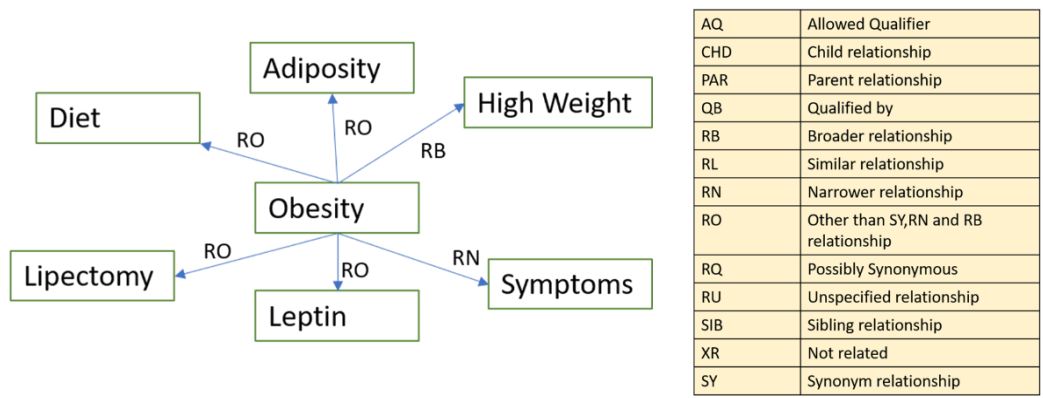

**Figure 2.** A Concept and its related concepts from UMLS Metathesaurus. UMLS relationships and their description.

Datasets

**Table 1.** An overview of the biomedical concept relatedness datasets. Here MW represents multi-word.

| Dataset | Source | Type | #concept Pairs | Annotation Scale | % MW concepts |
|---|---|---|---|---|---|
| MayoSRS [8] | UMLS | Hand-picked | 101 | 1-10 | 44% |
| UMNSRS-Rel [59] | UMLS | Hand-picked | 587 | 0-1600 | 2% |
| UMNSRS-Sim [59] | UMLS | Hand-picked | 566 | 0-1600 | 2% |
| EHR-RelB [6] | EHR | Co-occurring | 3630 | 0-3 | 89% |

We evaluated our approach on all four publicly available datasets including the recently released EHR-RelB dataset. A summary of these datasets is given in Table 1. EHR-RelB dataset is the largest publicly available semantic relatedness dataset having concept pairs that frequently co-occur in EHRs, unlike the other datasets which contain hand-picked concept pairs. EHR-RelB dataset with the largest number of concept pairs and having 89% of multi-word terms is the most challenging dataset compared to other existing datasets [6]. After excluding obsolete concepts (concepts not available in

UMLS 2019 AB version), there were 99 concepts in MayoSRS, 564 concepts in UMNSRS-Rel, and 543 concepts in UMNSRS-Sim. In EHR-RelB, only 3225 concept pairs were assigned with CUIs. So, we evaluated our approach on 3225 concept pairs in EHR-RelB.

**Methods**

Generation of Concept Vectors

To generate concept vectors, we encoded concept preferred terms using domain-specific models based on ELMo, BERT, and SBERT architectures. We used BioELMo and ClinicalELMo which are publicly available. ELMo is based on a two-layered BiLSTM model trained using language modeling objective. Initially, character-level word representations are generated using CharCNN and highway network layer. With these character-level word representations as input, two-layered BiLSTM sequentially processes embeddings of each word. We averaged the vector of all the words in the preferred term to generate the concept vector.

We used both Ontology Knowledge Free (OKF) models like PubMedBERT, BioBERT and BioClinicalBERT, Ontology Knowledge Injected (OKI) models like SapBERT, CoderBERT, UmlsBERT, and KbBERT. All these models are publicly available. In general, pre-trained language models like BERT can be used in two ways. The first one is to fine-tune the model using task-specific training instances after including task-specific layers. In this approach, weights of pre-trained language model and task-specific layers are updated. The second one is to get the contextual token embeddings without including any task-specific layers and no fine-tuning. In this approach, the weights of the pre-trained language model are kept fixed. In this paper, we used the second approach as we want only the contextual token embeddings for the given concept. The concept preferred term is added with the special tokens '[CLS]' and '[SEP]' like '[CLS] squamous cell carcinoma [SEP]'. The output of the last transformer encoder layer contains the final vectors of all the tokens. We averaged the final vectors of all the tokens to get the concept vector.

In the case of Sentence BERT models, we initially experimented with publicly available off-the-shelf models. As the off-the-shelf Sentence BERT models released by the authors of SBERT [49] are based on models pre-trained on general corpora, these models contributed to poor results. As part of preliminary experiments, we trained Sentence BERT models based on domain-specific BERT models in four ways. First, we trained the models using full SNLI and STSb datasets. Second, we trained the models using only 25k SNLI instances and STSb. Third, we trained the models using MedNLI [60] and BIOSSES [61] datasets. Fourth, we trained the models using 25k SNLI instances + MedNLI and STSb datasets. Out of all the four, we got the best results when the models were trained using only 25k SNLI instances and STSb. The possible reasons for this are a) SNLI dataset consists of around 570k instances gathered from a general text corpus. As these instances are large in number and out-of-domain, training the models on the full SNLI dataset reduced the performance. b) Small sizes of MedNLI dataset (around 14k instances) and BioSSES (only 100 instances). Further, we experimented by training the models on more SNLI instances (50K, 100K, 150K) and STSb. As SNLI instances are from general corpus, this reduced the performance. Finally, we trained all the domain-specific Sentence BERT models using 25k SNLI instances and full STSb dataset.

Retrofitting Concept Vectors

Concept vectors generated using domain-specific models consist of only the knowledge available in the pre-training text corpora. However, the quality of these concept vectors can be further improved using abundant relationship knowledge from UMLS. To inject UMLS relationship knowledge into concept vectors, we used retrofitting algorithm [58]. Retrofitting algorithm expects two text files as input. Each line of the first text file should contain concept preferred term and its vector and each line of the second text file should contain concept preferred term and preferred terms of its related concepts. For each concept in all four datasets, we gathered related CUIs from MRREL file. Further, we gathered preferred terms of all the CUIs from MRCONSO file. We used the code provided by the authors of retrofitting algorithm and set the number of iterations to 10 (default value).

**Evaluation Measures**

Methods for computing relatedness scores are evaluated on how much the computed scores correlate with the mean annotated scores. Initially, we computed relatedness scores using cosine similarity for each of the concept pairs. After that, we computed the spearmen rank correlation using the relatedness scores and the mean annotated scores. We compared our performance with the performances reported by Schulz and Juric [62], Moa and Fung [20], and Schulz et al. [6] in 2020. Schulz and Juric [62] evaluated the performance of 13 publicly available state-of-the-art biomedical word embeddings including BioWordVec [63] on MayoSRS and UMNSRS datasets. Schulz et al. [6] created EHR-RelB dataset and tested the performance of 13 publicly available biomedical word embeddings. Mao and Fung [20] evaluated their hybrid measure based on biomedical word embeddings and graph embeddings on UMNSRS datasets.

**RESULTS**

We conducted experiments in four stages. First, we evaluated the performance of various ELMo and BERT models, and the results are listed in Table 2. Second, we trained and evaluated the performance of various domain-specific Sentence BERT models and the results are listed in Table 2. SapBERT-S model (Sentence BERT model based on SapBERT) achieved the best results and so we conducted further experiments with this model. Third, we evaluated the impact of retrofitting concept vectors generated by SapBERT-S model using related concepts from UMLS relationships, and the results are listed in Table 3. Fourth, we evaluated the impact of retrofitting concept vectors generated by SapBERT-S using concepts from more than one UMLS relationship and the results are listed in Table 4.

As listed in Table 2, in the case of ELMo and BERT models, SapBERT achieved the best results on MayoSRS and EHR-RelB datasets, CoderBERT achieved the best results on both UMNSRS datasets.

**Table 2.** Performance of ELMo, BERT and SBERT based models on four datasets.

| Relationship | MayoSRS | UMNSRS-Rel | UMNSRS-Sim | EHR-RelB |
|---|---|---|---|---|
| No retrofitting | 0.5616 | 0.5053 | 0.5850 | 0.5292 |
| AQ | 0.5615 | 0.5053 | 0.5850 | 0.5292 |
| SIB | 0.6261 | 0.5161 | 0.6024 | 0.5376 |
| PAR | 0.5951 | 0.5372 | 0.6209 | 0.5598 |
| RB | 0.5920 | 0.5258 | 0.6067 | 0.5580 |
| RL | 0.5833 | 0.4979 | 0.5700 | 0.5393 |
| RU | 0.5616 | 0.5053 | 0.5850 | 0.5292 |
| QB | 0.3783 | 0.3915 | 0.4903 | 0.4948 |
| XR | 0.5615 | 0.5053 | 0.5850 | 0.5292 |
| CHD | **0.6266** | 0.5427 | 0.6209 | 0.5278 |
| SY | 0.5714 | 0.5258 | 0.5990 | 0.5345 |
| RQ | 0.6151 | 0.5696 | 0.6345 | **0.5874** |
| RN | 0.6093 | **0.5780** | **0.6500** | 0.5528 |
| RO | 0.6034 | 0.5459 | 0.6142 | 0.5535 |

BioELMo got better results compared to all OKF BERT models on all four datasets and it is illustrated in Figure 3. Among OKF BERT models, PubMedBERT model outperformed others on MayoSRS and EHR-RelB, while BioBERT outperformed others on UMNSRS-Rel and UMNSRS-Sim datasets.

**Table 3.** Performance of SapBERT-S concept vectors retrofitted using concepts from UMLS relationships.

| Model | MayoSRS | UMNSRS-Rel | UMNSRS-Sim | EHR-RelB |
|---|---|---|---|---|
| ELMo and BERT models | | | | |
| BiomedicalELMo | 0.4639 | 0.3957 | 0.4789 | 0.3989 |
| ClincalELMo | 0.3339 | 0.2841 | 0.3764 | 0.4066 |
| PubMedBERT | 0.4007 | 0.2658 | 0.3159 | 0.3676 |
| SapBERT | **0.5699** | 0.4532 | 0.5214 | **0.5253** |
| BioBERT | 0.3645 | 0.2775 | 0.3598 | 0.3607 |
| CoderBERT | 0.5205 | **0.4731** | **0.5432** | 0.5161 |
| KbBERT | 0.3352 | 0.2824 | 0.3774 | 0.4487 |
| BioclincialBERT | 0.1716 | 0.1627 | 0.2707 | 0.3700 |
| UmlsBERT | 0.1530 | 0.1949 | 0.3 | 0.3305 |
| Sentence BERT models | | | | |
| PubMedBERT-S | **0.5684** | 0.4354 | 0.5233 | 0.5260 |
| SapBERT-S | 0.5616 | **0.5053** | **0.5850** | **0.5292** |
| BioBERT-S | 0.4907 | 0.4086 | 0.4749 | 0.5015 |
| CoderBERT-S | 0.4721 | 0.4320 | 0.5067 | 0.4968 |
| KbBERT-S | 0.4570 | 0.4590 | 0.5218 | 0.4984 |
| BioclinicalBERT-S | 0.4127 | 0.4056 | 0.4645 | 0.4751 |
| UmlsBERT-S | 0.3449 | 0.3848 | 0.4360 | 0.4566 |

BioClincalBERT got the least scores on all four datasets.

**Table 4.** Performance of SapBERT-S concept vectors when retrofitted using concepts from multiple UMLS relationships.

| Relationships | MayoSRS | UMNSRS-Rel | UMNSRS-Sim | EHR-RelB |
|---|---|---|---|---|
| RQ+RN | 0.6506 | 0.5858 | 0.6594 | 0.5791 |
| RN+RO | 0.6070 | 0.5500 | 0.6178 | 0.5563 |
| RQ+RO | 0.6157 | 0.5566 | 0.6245 | 0.5610 |
| RQ+SY | 0.6169 | 0.5673 | 0.6583 | **0.5855** |
| RQ+PAR | 0.6027 | 0.5662 | 0.6446 | 0.5848 |
| RN+SY | 0.6147 | 0.5601 | 0.6457 | 0.5568 |
| RN+PAR | 0.6124 | 0.5648 | 0.6458 | 0.5688 |
| RQ+RN+RO | 0.6281 | 0.5599 | 0.6375 | 0.5624 |
| RN+SY+PAR | 0.6083 | 0.5641 | 0.6481 | 0.5672 |
| RQ+RN+SY | **0.6529** | **0.6002** | **0.6710** | 0.5800 |
| RQ+RN+PAR | 0.6171 | 0.5824 | 0.6479 | 0.5843 |
| RQ+RN+RL | 0.6502 | 0.5852 | 0.6488 | 0.5800 |
| RQ+SY+PAR | 0.6012 | 0.5775 | 0.6597 | 0.5828 |
| RQ+PAR+CHD | 0.6369 | 0.5661 | 0.6468 | 0.5639 |
| RQ+RN+RL+SY | 0.6505 | 0.5864 | 0.6505 | 0.5807 |
| RQ+RN+SY+PAR | 0.6160 | 0.5852 | 0.6527 | 0.5826 |
| RQ+RL+SY+PAR | 0.6025 | 0.5776 | 0.6496 | 0.5829 |
| RQ+RN+RO+SY | 0.6277 | 0.5614 | 0.6399 | 0.5703 |
| RQ+RN+RO+SY+PAR | 0.6284 | 0.5652 | 0.6418 | 0.5795 |
| RQ+RN+RO+SY+PAR+CHD | 0.6347 | 0.5650 | 0.6419 | 0.5759 |

As listed in Table 2, Sentence BERT models outperformed corresponding BERT models on all four datasets in most of the cases. As listed in Table 3, retrofitting using UMLS relationships improved scores in most of the cases except AQ, RU, XR, and QB. Among all UMLS relationships, retrofitting using a) CHD got the best results for MayoSRS b) RN got the best results for both UMNSRS datasets, and c) RQ got the best results for EHR-RelB. As listed in Table 4, the combination RQ+RN+SY got the best results for MayoSRS, UMNSRS-Rel, and UMNSRS-Sim datasets. In the case of EHR-RelB, any of the combinations could not improve the score compared to RQ relationship.

**DISCUSSION**

Here, we proposed a simple and novel approach for the automatic computation of semantic relatedness scores between biomedical concepts. Our approach being hybrid leverages the advantages in both knowledge-based and distributional approaches. Our approach utilizes publicly available BERT-based models, SNLI, and STSb datasets for training domain-specific Sentence BERT models and UMLS for expert curated domain knowledge.

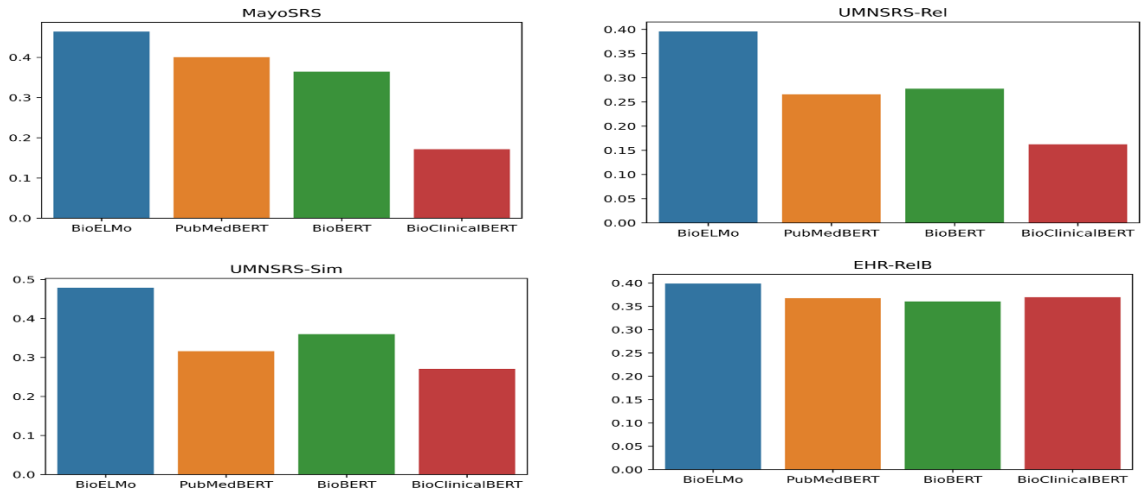

**Figure 3.** Comparison of BioELMo and OKF BERT (PubMedBERT, BioBERT, and BioClinicalBERT) models.

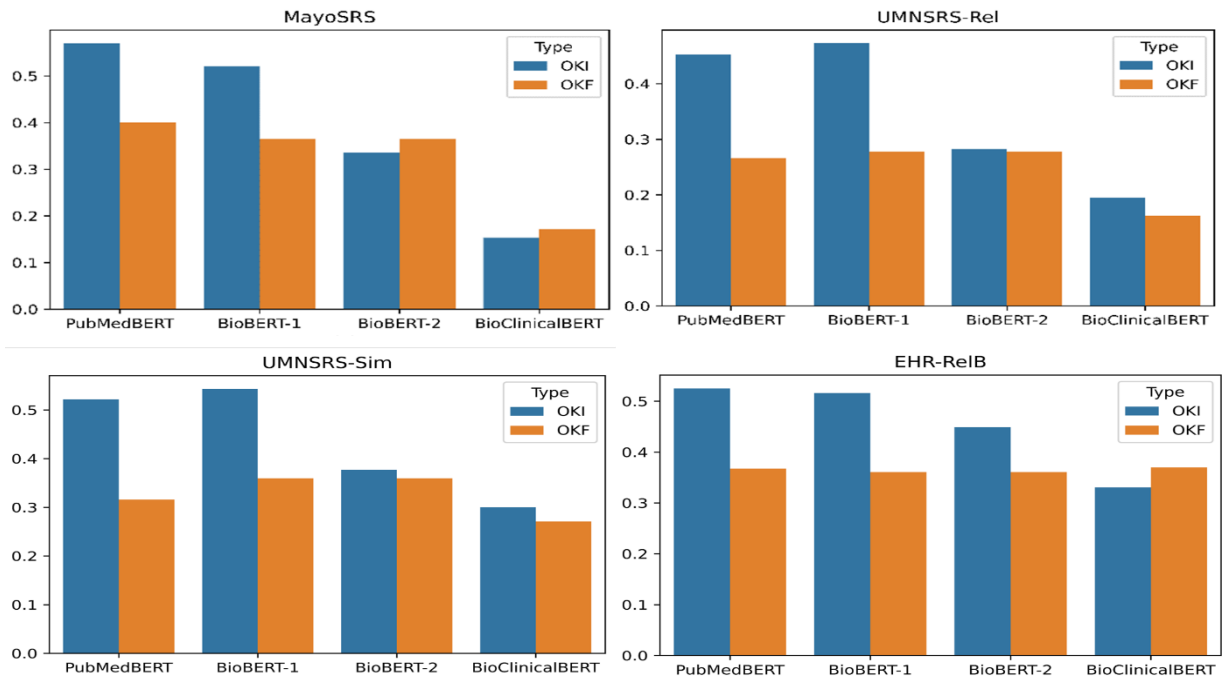

**Figure 4.** Comparison of OKI and OKF BERT models. SapBERT, CoderBERT, KbBERT and UmlsBERT are OKI versions of PubMedBERT, BioBERT, BioBERT and BioClinicalBERT respectively. BioBERT-1 represents CoderBERT and BioBERT, BioBERT-2 represents KbBERT and BioBERT.

As reported in Table 2, OKI BERT models contributed to better results compared to OKF models on all four datasets in most of the cases and it is illustrated in Figure 4. As reported in Table 3, Sentence

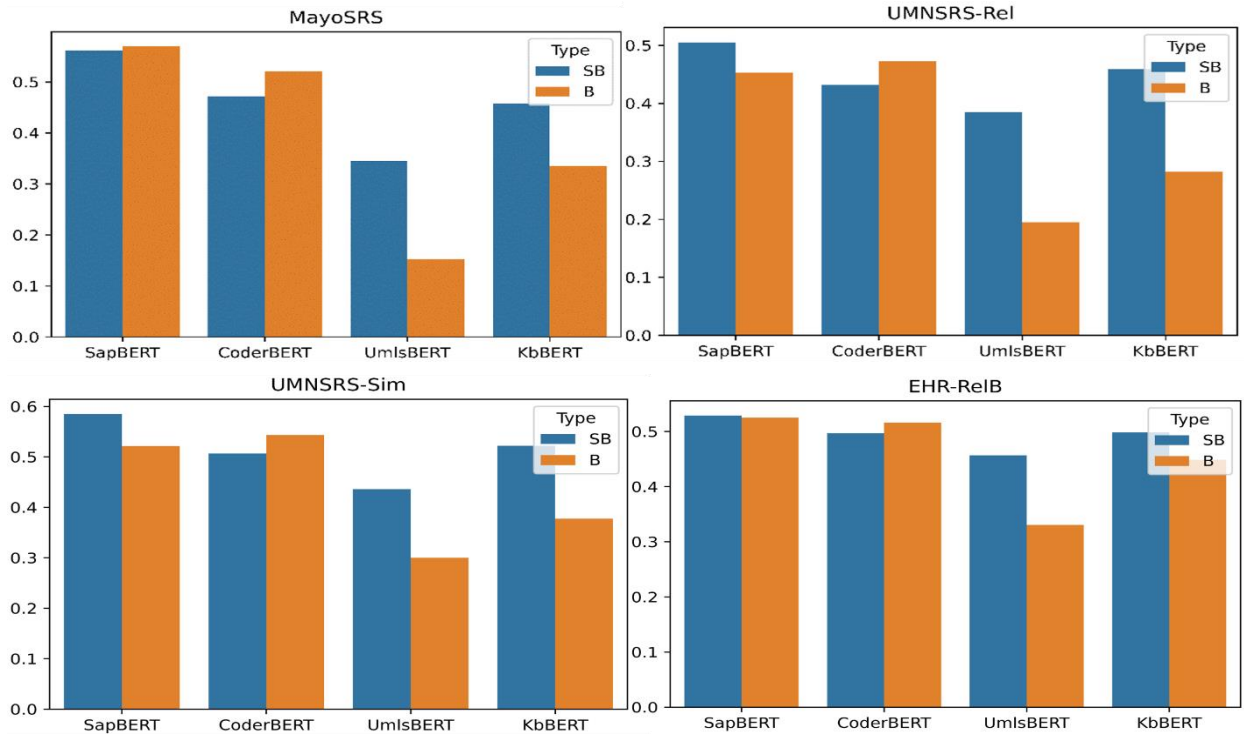

**Figure 5.** Comparison of OKI Sentence BERT and BERT models (OKI - Ontology Knowledge Injected, SB – Sentence BERT, B - BERT.

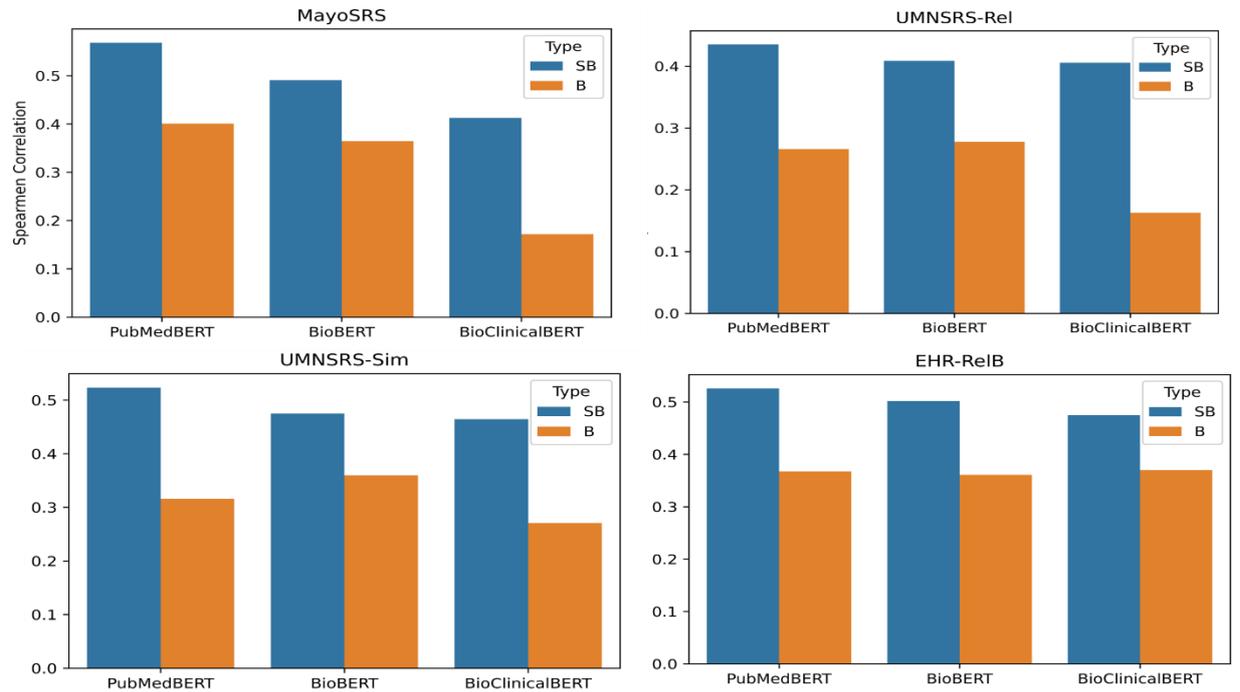

**Figure 6.** Comparison of OKF Sentence BERT and BERT models (OKF - Ontology Knowledge Free, SB – Sentence BERT, B - BERT).

BERT models (except in the case of CoderBERT) contributed to better results compared to corresponding BERT models on all four datasets. This shows that training BERT models using SNLI and

STSb datasets allow the models to learn more semantic information at the phrase or sentence level. As illustrated in Figures 5 and 6, improvements in scores were more in the case of OKF BERT models compared to OKI BERT models.

From the results presented in Table 3, we infer that enriching concept vectors with relationship knowledge from UMLS leads to better results except in a few cases. Retrofitting using relations like AQ, XR, RU resulted in the same performance as concepts vectors without retrofitting. It is because there were no related concepts from these relationships for any of the concepts in the datasets. This shows that the choice of relations in UMLS is crucial to get better scores. From Table 4, we observe that retrofitting using concepts from more than one UMLS relation improves the results except in the case of EHR-RelB. For example, the combination RQ+RN+SY achieved the best results on MayoSRS, and both UMNSRS datasets. We achieved spearmen rank correlation scores of 0.6529, 0.6002, 0.6710, and 0.5874 on MayoSRS, UMNSRS-Rel, UMNSRS-Sim, and EHR-RelB datasets respectively. Schulz et al. [6] achieved correlation scores of 0.49 on EHR-RelB and Schulz and Juric [62] achieved scores of 0.57, 0.59, and 0.66 on MayoSRS, UMNSRS-Rel, and UMNSRS-Sim datasets. Mao and Fung [20] achieved correlation scores of 0.5904 and 0.6288 on UMNSRS-Rel and UMNSRS-Sim datasets. Our approach achieved the best results on all four datasets. Our approach achieved improvements of 0.8290, 0.1002, 0.1110, and 0.9740 on MayoSRS, UMNSRS-Rel, UMNSRS-Sim, and EHR-RelB datasets respectively compared to existing approaches. MayoSRS includes 44% of multi-word concepts and EHR-RelB includes 89% of multi-word concepts, but the percentage of multi-word concepts is only 2% in UMNSRS datasets. Sentence BERT models learn more semantic information at the phrase or sentence level due to training on SNLI and STSb datasets and can represent multi-word concepts better compared to word embeddings. So, our approach achieved more improvements in the case of MayoSRS and EHR-RelB datasets compared to both UMNSRS datasets.

Our approach has several advantages compared to the existing approaches. First, for a given concept our approach can leverage related concepts from all the available relations as well as from all the vocabularies in UMLS as there is no restriction on the number of related concepts in retrofitting algorithm. However, the existing hybrid approach [20] leveraged UMLS relationship knowledge in the form of knowledge graph embeddings generated using only Parent-Child relationships gathered from only two vocabularies, SNOMED-CT [64] and MedDRA [65]. There are two limitations in leveraging UMLS relationship knowledge using knowledge graph embeddings. They are a) in the case of Knowledge Graph Embedding algorithms, there is a limit on the number of nodes a graph can have [20] and so it is not possible to include related concepts from more relationship types as well as related concepts from more vocabularies and b) it is not guaranteed that the related concepts are always linked in ontologies which limits the quality of embeddings generated and hence results in poor performance [20]. So, retrofitting is a much better alternative to leverage relationship knowledge compared to knowledge graph embeddings. Second, our approach can be used to compute relatedness or similarity scores between any UMLS concepts, unlike the approaches based on path information, information content measures, or CUI embeddings. Third, our approach is easily scalable to new concepts as computing concept vectors using Sentence BERT model and retrofitting takes very less time, unlike graph embeddings methods which are to be re-trained from scratch after adding new concepts in the graph. Fourth, our approach can better represent multi-word concepts as it uses the Sentence BERT model.

There are some limitations to our study. We leveraged only concept preferred term and relations from UMLS. We experimented with only retrofitting algorithm to inject relationship knowledge into concept vectors. As future work, we want to leverage other information from UMLS like semantic type and concept definitions. We also want to experiment with other algorithms like Extrofitting [66] to inject ontology knowledge into concept vectors.

**CONCLUSION**

Sentence BERT models trained on SNLI and STSb datasets can better represent multi-word concepts. Injecting UMLS relationship knowledge into concept vectors using retrofitting algorithm improves relatedness scores. Our semantic relatedness measure based on Sentence BERT and retrofitting algorithm achieved the best results on four publicly available datasets including the largest publicly available EHR-RelB dataset.

**COMPETING INTERESTS**

No competing interests